\documentclass[letterpaper, 10 pt, journal, twoside]{IEEEconf}

\usepackage{style}
\begin{document}
\maketitle
\thispagestyle{empty}

\begin{abstract}
The key innovation of our analytical method, CaRT, lies in establishing a new hierarchical, distributed architecture to guarantee the safety and robustness of a given learning-based motion planning policy. First, in a nominal setting, the analytical form of our CaRT safety filter formally ensures safe maneuvers of nonlinear multi-agent systems, optimally with minimal deviation from the learning-based policy. Second, in off-nominal settings, the analytical form of our CaRT robust filter optimally tracks the certified safe trajectory, generated by the previous layer in the hierarchy, the CaRT safety filter. We show using contraction theory that CaRT guarantees safety and the exponential boundedness of the trajectory tracking error, even under the presence of deterministic and stochastic disturbance. Also, the hierarchical nature of CaRT enables enhancing its robustness for safety just by its superior tracking to the certified safe trajectory, thereby making it suitable for off-nominal scenarios with large disturbances. This is a major distinction from conventional safety function-driven approaches, where the robustness originates from the stability of a safe set, which could pull the system over-conservatively to the interior of the safe set. Our log-barrier formulation in CaRT allows for its distributed implementation in multi-agent settings. We demonstrate the effectiveness of CaRT in several examples of nonlinear motion planning and control problems, including optimal, multi-spacecraft reconfiguration.
\end{abstract}
\section{Introduction}
\label{sec_introduction}
Learning-based control has been a subject of intense study for solving large-scale and complex problems that conventional approaches fail to handle; one example of which is to construct a real-time, nonlinear motion planning and control algorithm for multi-agent robotic and aerospace systems. 
The safety and robustness of the machine-learning approaches, however, highly depend on a number of factors such as structures of approximation models, tools used to generate training and test data, data pre-conditioning, learning duration, and training schemes. This makes it difficult to obtain a universal result applicable to any learning approach, treating some parts of the performance as a black box. 

The purpose of this paper is to provide control theoretical safety and robust tracking guarantees to given learned motion planning policies for nonlinear multi-agent systems, independently of the performance of the learning approaches used in designing the learned policy. Our approach called CaRT (Certified Safety and Robust Tracking) is two-fold as follows.
\subsubsection*{Contributions}
First, assuming that the dynamics model is free of disturbance and uncertainty, we construct an optimal safety filter for generating a safe target trajectory of nonlinear multi-agent systems in this ideal setting. We explicitly derive an analytical form of the optimal control input processed by the safety filter, which minimizes its deviation from the control input of a given learned motion planning policy while ensuring safety. We utilize the log-barrier formulation~\cite{citeulike:163662} so that the global safety violation can be decomposed as the sum of the local safety violations, allowing for the distributed implementation of our analytical safety filter in a multi-agent setting. 

Second, we hierarchically design a robust filter based on contraction theory~\cite{Ref:contraction1,tutorial} for guaranteeing exponential tracking of the safe target trajectory, even when the nominal dynamics model is subject to unknown deterministic and stochastic perturbation. This is because, although the safety filter mentioned earlier (including the CLF-CBF control~\cite{cbf_clf}, see Sec.~\ref{sec_clfcbf}) also possesses a robustness guarantee, it is based primarily on the repelling force of its control input to push the system state back to the safe set (\ie{}, asymptotic/exponential stability of the safe set). This can pull the system over-conservatively to the interior of the safe set, \eg{}, when implementing the filter in real-world systems involving discretization of the control policy and dynamics. Adding a robust filter hierarchically, based on incremental stability of system trajectories, explicitly separates disturbances from safety violations and unknown dynamics, thus alleviating the burden of the safety filter in robustly dealing with the disturbance (see~Sec.~\ref{sec_high_level_overview} and~\ref{sec_built_in_robustness} for details). This filter also provides an analytical form of the control input for real-time implementation. The application of these two filters in CaRT results in provable safety and robust tracking guarantees augmented on top of the given learned motion planning policy, without prior knowledge of its original learning performance. Note that the analytical solution processed by these filters is still differentiable, thereby allowing for the end-to-end learning of the safe control policy as in~\cite{glas}. 

As for the nominal dynamics model, we start our discussion with the Lagrangian dynamical systems~\cite[p. 392]{Ref_Slotine}, which can be used to describe a wide range of robotic and aerospace systems. We then extend the safety and robustness results to general control-affine nonlinear systems. Our approach is demonstrated in several nonlinear multi-agent motion planning and control problems.
\subsubsection*{Related Work}
The high-level comparison between our approach, CaRT, and the methods outlined in this section will be revisited in Sec.~\ref{sec_high_level_overview} in more detail. 

The nonlinear robustness and stability can be analyzed using a Lyapunov function, which gives a finite tracking error with respect to a given target trajectory under the presence of external disturbances, including approximation errors of given learned motion planning policies. 
This framework thus provides a one way to ensure safety and robustness in learning-based control methods (see, \eg{}, \cite{tutorial,8967820,chuchu_survey} and references therein), which depends on the knowledge and the size of the approximation error of a given learned motion planning policy. Such information could be conservative for previously unseen data or available only empirically.

Control barrier functions, in contrast, guarantee the safety of nonlinear systems in real time without any knowledge of the learned motion planning errors at all. Formulating a CLF-CBF Quadratic Program (QP)~\cite{cbf_clf,robust_cbf,high_order_cbf1} also gives some guarantees on robustness. As shall be elaborated in Sec.~\ref{sec_clfcbf}, such guarantees are based on the repelling force of its control input to exponentially/asymptotically push the system state back to the stable safe set, which could lead to unnecessarily large/conservative control inputs for safety.

The purpose of our work is to propose a hierarchical approach to combine the best of both of these methods for safety and robustness, by performing contraction theory-based robust tracking of a provably safe trajectory, generated by a safety filter with the learned policy. The additional tracking-based robust filter is for reducing the burden in dealing with disturbances (see Sec.~\ref{sec_high_level_overview} and~\ref{sec_built_in_robustness}).
\subsubsection*{Notation}
\label{notation}
For $A \in \mathbb{R}^{n \times n}$, we use $A \succ 0$, $A \succeq 0$, $A \prec 0$, and $A \preceq 0$ for the positive definite, positive semi-definite, negative definite, negative semi-definite matrices, respectively. For $x \in \mathbb{R}^n$ and $A \in \mathbb{R}^{n \times m}$, we let $\|x\|$, $\|x\|_{\Xi}$, $\|A\|$, $\|A\|_F$ denote the Euclidean norm, weighted $2$-norm (\ie{}, $\|x\|_{\Xi}=\sqrt{x^{\top}\Xi x}$ for $\Xi \succ 0$), induced 2-norm, and Frobenius norm, respectively. Also, $\mathop{\mathbb{E}}$ denotes the expected value operator.
\section{Problem Formulation}
\label{sec_problem}
We consider the following multi-agent Lagrangian dynamical system, perturbed by deterministic disturbance $d^{i}(x,t)$ with $\sup_{x,t}\|d^{i}(x,t)\|=\bar{d}^{i}\in[0,\infty)$ and Gaussian white noise of a Wiener process $\mathscr{W}(t)$ with $\sup_{x,t}\|\Gamma^{i}(x,t)\|_F = \bar{\gamma}^{i}\in[0,\infty)$:
\begin{align}
    \label{eqlagrange}
    &M^{i}(p^{i})dv^{i}+(C^{i}(p^{i},v^{i})v^{i}+G^{i}(p^{i})+D^{i}(p^{i},v^{i}))dt \\
    &= ({u}^{i}+d^{i}(x^{i},t))dt+\Gamma^{i}(x^{i},t)d\mathscr{W}^{i}(t),~i=1,\cdots,N
\end{align}
where $t \in \mathbb{R}_{+}$, $x^{i} = [{p^{i}}^{\top},{v^{i}}^{\top}]^{\top}$, $i$ is the index of the $i$th agent, $p^{i}:\mathbb{R}_{+} \mapsto \mathbb{R}^{n}$ and $v^{i}:\mathbb{R}_{+} \mapsto \mathbb{R}^{n}$ 
are the generalized position and velocity of the $i$th agent ($\dot{p}^{i}=v^{i}$), ${u}^{i} \in \mathbb{R}^m$ ($m=n$ in this case) is the system control input, $M^{i}$, $C^{i}$, $G^{i}$, and $D^{i}$ are known smooth functions that define the Lagrangian system, $d^{i}$ and $\Gamma^{i}$ are unknown bounded functions for external disturbances, $\mathscr{W}^{i}$ is a $w$-dimensional Wiener process, and we consider the case where $\bar{d}^{i},\bar{\gamma}^{i} \in [0,\infty)$ are given. We have $M^{i}(p^{i})\succ 0$ and that the matrix $C^{i}(p^{i},\dot{p}^{i})$ is selected to make $\dot{M}^{i}(p^{i})-2C^{i}(p^{i},v^{i})$ skew-symmetric. Hence, $z^{\top}(\dot{M}^{i}(p^{i})-2C^{i}(p^{i},v^{i}))z=0,~\forall z\in\mathbb{R}^n$~\cite[p. 392]{Ref_Slotine}. We also consider the following general control-affine nonlinear system:
\begin{align}
    \label{equnderactuated}
    dv^{i} =& (f^{i}(p^{i},v^{i},t)+B(p^{i},v^{i},t)u^{i})dt \\
    &+d^{i}(x^{i},t)dt+\Gamma^{i}(x^{i},t)d\mathscr{W}(t),~i=1,\cdots,N
\end{align}
where $f^{i}$, and $B^{i}$ are known smooth functions, $u^{i} \in \mathbb{R}^m$ is the system control input, and the other notations are consistent with that of~\eqref {eqlagrange}. We assume the existence and uniqueness conditions of the solutions of~\eqref{eqlagrange} and~\eqref{equnderactuated} as in~\cite[p. 105]{arnold_SDE}.

The nonlinear motion planning problem of our interest is defined as follows:
\begin{align}
    &u_{\rm global}(t) = \text{arg} \min_{\{u^{i}(t)\in\mathbb{R}^m\}_{i=1}^N} \int^{t_f}_0 c(x(\tau),u(\tau),\tau)d\tau \label{def_motion_planning} \\
    &\text{\st{}~\eqref{eqlagrange} or \eqref{equnderactuated} with $d^{i}=0$ and $\Gamma^{i}=0$},~x^{i}(t_f)=x_f^{i},~x^{i}(0)=x^{i}_0 \nonumber \\
    &\text{\textcolor{white}{\st{}~}}\|p^{i}(t)-p^{j}(t)\| \geq r_s,~\forall t, ~\forall i,~j\neq i
\end{align}
where $x(t)=\{x^{i}(t)\}_{i=1}^N$, $u(t)=\{u^{i}(t)\}_{i=1}^N$, $c(x(t),u(t),t)$ is a user-specified cost at time $t$, $t_f$ is the terminal time, $x_0^{i}$ and $x^{i}_f$ are the initial and terminal state, respectively, $r_s$ is the minimal safe distance between $i$th agent and other objects, and $j$ is the index denoting other agents and obstacles. The control policy $u_{\rm global}$ generates the reference trajectory $x_{\rm global}$ of Fig.~\ref{concept_fig1} to be discussed in this section.
\subsection{Learned Distributed Motion Planning Policy}
\label{sec_learned_policy}
Let $\mathscr{N}$ denote the set of all the $N$ agents, $\mathscr{M}$ denote the set of all the $M$ static obstacles, and $o^{i}$ denote the local observation of the $i$th agent given as follows: 
\begin{align}
    \label{observation}
    o^{i} =& (x^{i},\{x^{j}\}_{j\in\mathscr{N}^{i}},\{p^{j}\}_{j\in\mathscr{M}^{i}})
\end{align}
where $x^{i}$ is the state of the $i$th agent, $\{x^{j}\}_{j\in\mathscr{N}^{i}}$ are the states of neighboring agents defined with $\mathscr{N}^{i} = \{j\in\mathscr{N}|\|p^{i}-p^{j}\| \leq r_{\mathrm{sen}}\}$, $\{p^{j}\}_{j\in\mathscr{M}^{i}}$ are the positions of neighboring static obstacles defined with $\mathscr{M}^{i} = \{j\in\mathscr{M}\ |\ \|p^{i}-p^{j}\| \leq r_{\mathrm{sen}}\}$, and $r_{\mathrm{sen}}$ is the sensing radius. 

In this paper, we assume that we have access to a learned distributed motion planning policy ${u}_{\ell}^{i}(o^{i})$ obtained by~\cite{glas}. In particular,
(i) we generate demonstration trajectories by solving the global nonlinear motion planning~\eqref{def_motion_planning} and
(ii) extract local observations from them for deep imitation learning to construct ${u}_{\ell}^{i}(o^{i})$. Our differentiable safety and robust filters to be seen in Sec.~\ref{sec_safety} and~\ref{sec_robustness} can be used also in this phase to allow for end-to-end policy training as in~\cite{glas}.
\subsection{Augmenting Learned Policy with Safety and Robustness}
\label{sec_high_level_overview}
As discussed in Sec.~\ref{sec_introduction}, directly using the learned motion planning policy has the following two issues in practice: 
(i) even in nominal settings without external disturbance in~\eqref{eqlagrange} and~\eqref{equnderactuated}, the system solution trajectories computed with the learned motion planning policy could violate safety requirements due to learning errors, and
(ii) the learned policy lacks formal mathematical guarantees of safety and robustness under the presence of external disturbance.
Before going into details, let us see how we address these two problems analytically in real-time for the general systems~\eqref{eqlagrange} and~\eqref{equnderactuated}, optimally and independently of the performance of the learning method used in the learned motion planning policy. 
\subsubsection{CaRT Safety Filter and Built-in Robustness}
\begin{figure}
    \centering
    \includegraphics[width=80mm]{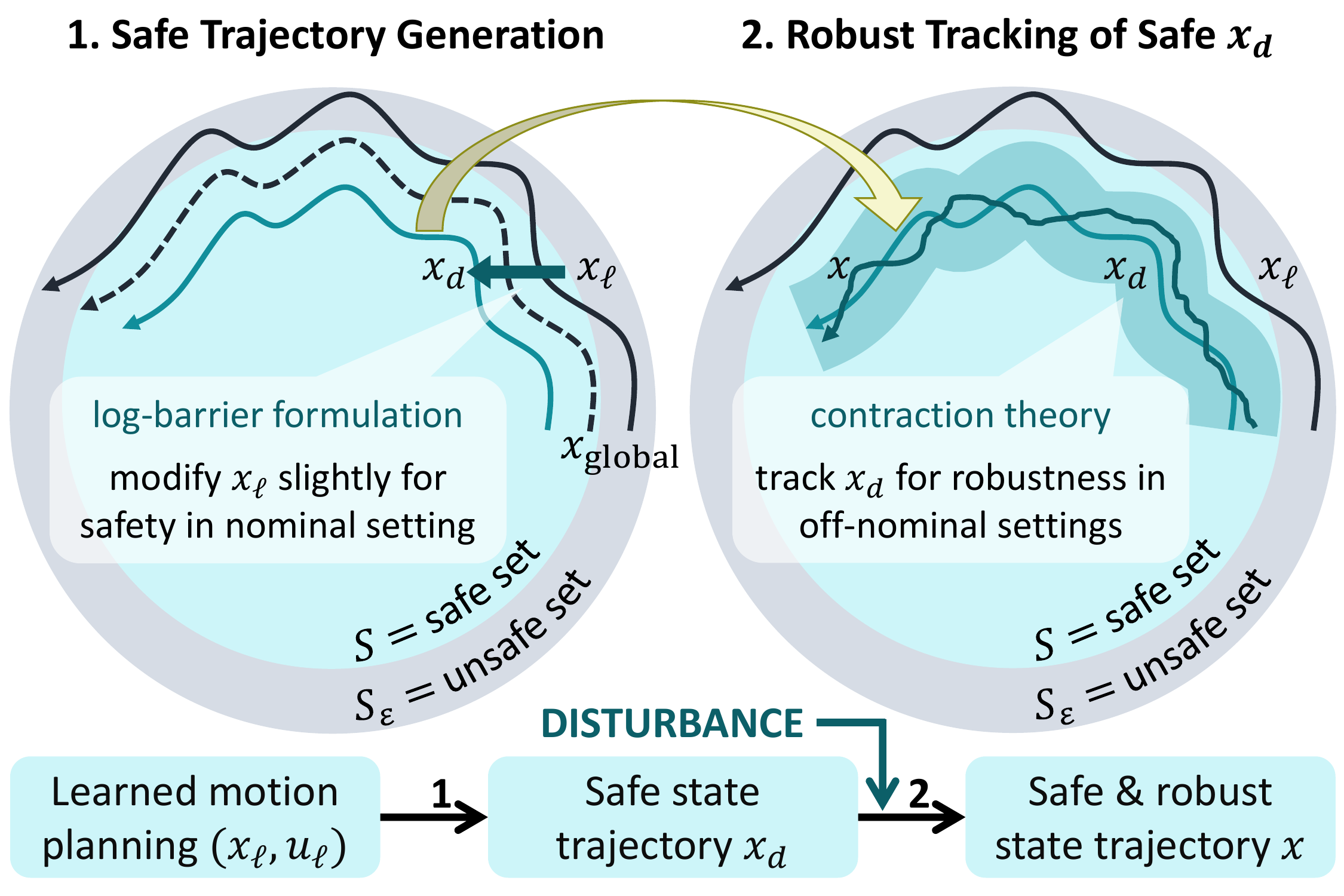}
    \vspace{-1.0em}
    \caption{Conceptual illustration of CaRT for a single agent, showing the hierarchical combination of our safety filter and robust filter, where $S$ is the safe set, $S_{\varepsilon}$ is some fictitious unsafe set containing learned trajectory $x_{\ell}$ with learning error $\varepsilon>0$, $x_{\rm global}$ is a reference trajectory given by global motion planner~(see~(\ref{def_motion_planning})), $x_{d}$ is CaRT's target safe trajectory, and $x$ is CaRT's actual state trajectory subject to disturbance. Note that we use a log-barrier formulation for the safe trajectory generation, which allows for the distributed and analytical implementation of CaRT.}
    \vspace{-1.0em}
    \label{concept_fig1}
\end{figure}
\begin{figure}
    \centering
    \includegraphics[width=80mm]{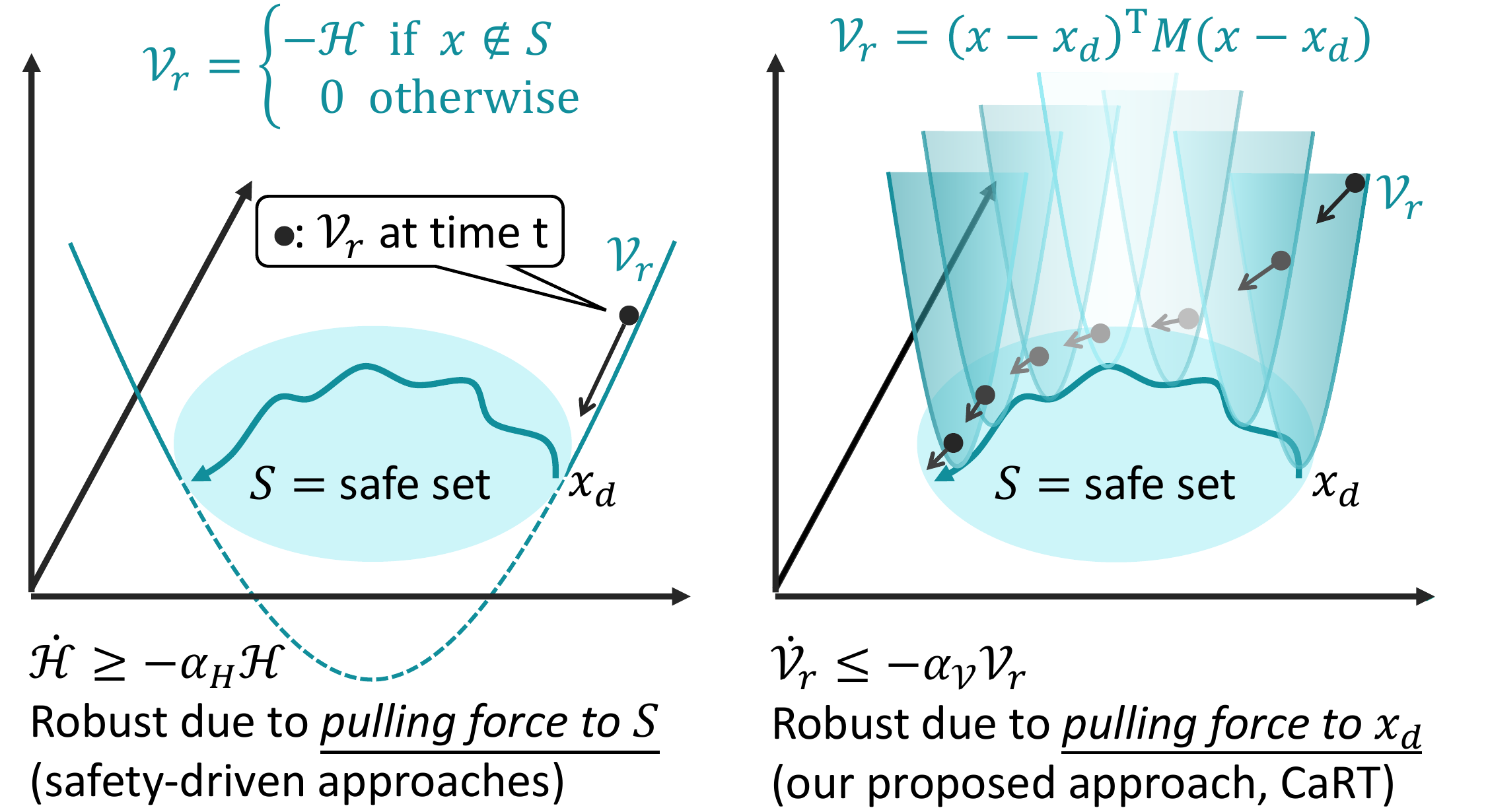}
    \vspace{-1.0em}
    \caption{Different sources of robustness in conventional safety-driven approaches (left, \eg{}, CLF-CBF~\cite{cbf_clf}) and CaRT (right), where $\alpha_h,\alpha_{\mathcal{V}}>0$, $\mathcal{H}$ is some safety function, $S$ is its safe set, $\mathcal{V}_r$ are respective Lyapunov functions for robustness, $x_d$ is a target trajectory, and $M$ is a contraction metric~\cite{Ref:contraction1,tutorial}. Safety-driven approaches are robust due to the stability of the safe set, while CaRT is robust due to the incremental stability of the closed-loop system with respect to the safe target trajectory $x_d$ (see Sec.~\ref{sec_high_level_overview}).}
\vspace{-1em}
    \label{concept_fig2}
\end{figure}
Given the learned policy ${u}_{\ell}^{i}$, we slightly modify it using our CaRT distributed safety filter to ensure the agents' safe operation. This is achieved by imposing a log-barrier safety constraint to guarantee safety under the presence of learning error $\varepsilon >0$, as in the left-hand side of Fig.~\ref{concept_fig1}. Intuitively, since the learning error $\varepsilon$ is expected to be small empirically, the contribution required for the safety filter is also expected to be small in a \textit{nominal} setting in practice. Note that we use the log-barrier formulation for the safe of real-time, distributed implementation of the safety filter in nonlinear multi-agent settings.

However, the robustness of such safety-driven approaches results from the stability of the safe set $S$~\cite{robust_cbf}, as shown in the left-hand side of Fig.~\ref{concept_fig2}. This yields the pulling force to the set $S$ that could be undesirably large in \textit{off-nominal} settings, which could then lead to an unnecessarily large tracking error. This is especially true, \eg{}, in real-world scenarios involving the discretization of the control policy and dynamics.
\subsubsection{CaRT Robust Filter and Tracking-based Robustness}
\label{sec_trade_off}
Instead of handling both safety and robustness just by the safety filter, we can further utilize our CaRT robust filter hierarchically to take over the role of disturbance attenuation in \textit{off-nominal} settings. As depicted in the right-hand side of Fig.~\ref{concept_fig1}, this is achieved by contraction theory-based robust tracking~\cite{Ref:contraction1,tutorial}, which guarantees the \textit{off-nominal} system state to stay in a bounded tube around the safe target trajectory $x_{d}$. Again, $x_d$ is a slight modification of the learned trajectory $x_{\ell}$ in a \textit{nominal} setting, processed through the CaRT safety filter.

We still use the Lyapunov formulation as in the safety filter for robustness, but now the Lyapunov function is defined incrementally as $\mathcal{V}_r = (x-x_{d})^{\top}M(x-x_{d})$, where $x$ is the off-nominal system state and $M\succ 0$ is the contraction metric~\cite{Ref:contraction1,tutorial}. As shown in the right-hand side of Fig.~\ref{concept_fig2}, improving the robustness performance here will simply result in superior tracking of the safe trajectory $x_{d}$, and thus can be achieved without losing too much information of the learned trajectory~${x}_{\ell}$. In other words, safety is handled directly with $x \in S$ in the robust filter, unlike safety-driven approaches with indirect derivative safety constraints $\dot{\mathcal{H}} \geq -\alpha_{\mathcal{H}} \mathcal{H}$.

These observations imply that 
\begin{enumerate}[label={\color{caltechgreen}{(\alph*)}}]
    \item when the learning error is much larger than the size of the external disturbance, then we can use our CaRT safety filter and its built-in robustness, and 
    \label{item_safety_robust}
    \item when the learning error is much smaller than the size of the external disturbance, which is often the case, then we can 1) modify the learned policy slightly with our CaRT safety filter in a \textit{nominal} setting, and 2) handle disturbance primarily with our CaRT tracking-based robust filter in \textit{off-nominal} settings, hierarchically on top of the safety filter. 
    \label{item_robust_robust}
\end{enumerate}
\subsubsection{Relationship with CLF-CBF}
\label{sec_clfcbf}
The CLF-CBF control~\cite{cbf_clf} also considers safety and robustness. In our context, it constructs an optimal control input by solving a QP to minimize its deviation from the learned motion planning policy, subject to the safety constraint and the stability constraint. Since safety is its priority, the stability constraint has to be relaxed to $\dot{\mathcal{V}}_r \leq -\alpha_\mathcal{V} \mathcal{V}_r+\rho$, where $\rho$ is for QP feasibility and the Lyapunov function $\mathcal{V}_r$ is now defined as $\mathcal{V} = (x-x_{\ell})^{\top}M(x-x_{\ell})$ for the learned trajectory $x_{\ell}$ of Fig.~\ref{concept_fig1}. We list key differences between the CLF-CBF controller and our approach, CaRT:
\begin{enumerate}[label={\color{caltechblue}{(\alph*)}}]
    \item The primary distinction is the direction in which the respective tracking component steers the system.
    The CLF-CBF stability component steers towards the learned trajectory, which, because of learning error, might not be safe. This means that tracking stability can be compromised for CLF-CBF by prioritizing safety over stability and robustness.  
    In contrast, CaRT's robust filter steers the system toward a certified safe trajectory, generated by the previous layer in the hierarchy, CaRT's safety filter.
    Because of this distinction, whereas CaRT can safely reject large disturbances with a large tracking gain $\alpha_{\mathcal V}$, this strategy is not practical for the CLF-CBF controller, which is forced to reject disturbances with large safety gain~$\alpha_{\mathcal H}$. This could pull the system over-conservatively towards the interior of the safety set as shown in Fig.~\ref{concept_fig2}. \label{clfcbf_item1}
    \item The secondary distinction is that CLF-CBF requires solving a QP with a given Lyapunov function, while CaRT provides an explicit way to construct the incremental Lyapunov function using contraction theory, and gives an analytical solution for the optimal control input in a distributed manner.
    This makes CaRT end-to-end trainable with a faster computation evaluation time. \label{clfcbf_item3}
\end{enumerate}


The trade-off of Sec.~\ref{sec_trade_off} and the strengths implied in Sec.~\ref{sec_clfcbf} will be demonstrated in Sec.~\ref{sec_simulation}.
\section{Analytical Form of Optimal Safety Filter}
\label{sec_safety}
In this section, we derive an analytical way to design a safety filter that guarantees safe operations of the systems~\eqref{eqlagrange} and~\eqref{equnderactuated} when $d^{i} = 0$ and $\Gamma^{i} = 0$. 
One of the benefits of the log-barrier formulation in the following is that the global safety violation can be decomposed as the sum of the local safety violations, allowing for the distributed implementation of our analytical safety filter in a multi-agent setting.

Although~\eqref{safety_def} considers collision-free operation as the objective of safety to show one example of its use, we remark that all the proofs to be discussed work also with general notions of safety with a slight modification with~\eqref{general_safety_func}. Collision avoidance is just one of the most critical safety requirements in a multi-agent setting.
\subsection{Multi-Agent Safety Certificates}
We use the following for local safety certificates:
\begin{align}
    \label{safety_def}
    \psi^{i}(o^{i}) = -\log \prod_{j\in \mathcal{S}^i}h(p^{ij}),~h(p^{ij}) = \frac{\|p_{ij}\|_{\Xi}-(r_{s}+\Delta r_{s})}{r_{\mathrm{sen}}-(r_{s}+\Delta r_{s})}
\end{align}
where $p^{ij} = p^{j}-p^{i}$, $\mathcal{S}^i$ is the set of all the neighboring objects (\ie{}, agents and obstacles), $r_{s}$ is the minimal safe distance between $i$th agent and other objects, $r_{\mathrm{sen}}$ is the sensing radius, and $\Delta r_{s} > 0$ is a positive scalar parameter to account for the external disturbance to be formally defined in Sec.~\ref{sec_robustness}. We use the weighted $2$-norm $\|\cdot\|_{\Xi}$ with the weight $\Xi\succ0$ to consider a collision boundary defined by an ellipsoid. The parameters are selected to satisfy $\sup_{\|p_{ij}\|\leq r_{\mathrm{sen}}}\|p_{ij}\|_{\Xi} \leq r_{\mathrm{sen}}\text{ and }r_{\mathrm{sen}}-(r_{s}+\Delta r_{s}) > 0$ to ensure $\psi^{i}(o^{i}) \geq 0$ always when the distance to the $j$th object is less than $r_{\mathrm{sen}}$. Having a negative value of $h$ implies a safety violation. The idea for our CaRT safety filter is first to construct a safe target velocity given as 
\begin{align}
    \label{safe_velocity}
    v^{i}_{d}(o^{i}) = -k_{p}\nabla_{p^{i}}\psi^{i}(o^{i})
\end{align}
where $k_{p} > 0$, and then track it optimally using the knowledge of Lagrangian systems and contraction theory.

When dealing with general safety $h(x^{i},x^{j}) \geq 0$ for each $i$ and $j$, we can also use the local safety function~\eqref{safety_def} modified as
\begin{align}
    \label{general_safety_func}
    \psi^{i}(o^{i}) = -\log \prod_{j\in \mathcal{S}^i}h(x^{i},x^{j}).
\end{align}
As mentioned earlier, our framework to be discussed can handle general notions of safety with just a slight modification using~\eqref{general_safety_func} instead of~\eqref{safety_def}.
\begin{remark}
\label{remark_safety_relax}
The velocity~\eqref{safe_velocity} renders the single integrator system (\ie{}, $\dot{p}^{i} = v^{i}$) safe~\cite{glas}. Note that instead of the condition $\dot{\psi}\leq 0$ in~\cite{glas,lagrangian_collision}, we could use $\dot{\psi}\leq \alpha({h})$ to increase the available set of control inputs~\cite{cbf_clf}, where $\psi$ is a barrier function, ${h}$ is a safety function associated with $\psi$, and $\alpha$ is a class $\mathcal{K}$ function~\cite[p. 144]{Khalil:1173048}. This requires an additional global Lipschitz assumption on $\alpha$ as seen in~\cite{high_order_cbf1}.
\end{remark}
\subsection{Optimal Safety Filter for Lagrangian Systems}
\label{sec_optimal_safety}
Given the learned motion planning policy ${u}_{\ell}^{i}(o^{i})$ of Sec.~\ref{sec_learned_policy} for the system~\eqref{eqlagrange}, we design a control policy ${u}_{d}^{i}$ processed by our CaRT safety filter as follows:
\begin{align}
    \label{safe_control}
    {u}_{d}^{i}(o^{i}) =  {u}_{\ell}^{i}(o^{i})-  \begin{cases} 
   \frac{e^{i}_v({u}_{\ell}^{i}(o^{i})-\bar{{u}}_{d}^{i})^{\top}e^{i}_v}{\|e^{i}_v\|^2} & \text{if $({u}_{\ell}^{i}(o^{i})-\bar{{u}}_{d}^{i})^{\top}e^{i}_v > 0$} \\ 
    0 & \text{otherwise}
    \end{cases}~~~~
\end{align}
where $e^{i}_v=v^{i}-v^{i}_{d}(o^{i})$ for $v^{i}_{d}(o^{i})$ of~\eqref{safe_velocity}, $\bar{{u}}_{d}^{i}$ is given as
\begin{align}
    \label{def_us_bar_lagrange}
    \bar{{u}}_{d}^{i}=M^{i}\dot{v}^{i}_{d}+C^{i}v^{i}_{d}+G^{i}+D^{i}+v^{i}_{d}-k_{v}M^{i}e^{i}_v
\end{align}
with its arguments omitted and $k_{p},k_{v}>0$ being design parameters. The controller~\eqref{safe_control} is well-defined even with the division by $\|e^{i}_v\|$ as the relation $({u}_{\ell}^{i}(o^{i})-\bar{{u}}_{d}^{i})^{\top}e^{i}_v = 0 \leq 0$ holds when $\|e^{i}_v\| = 0$.
We have the following for the safety guarantee.
\begin{theorem}
\label{thm_lagrangian_safe}
Consider the following optimization problem:
\begin{align}
    \label{eqopt_lag}
    &{u}^{i}_{\mathrm{opt}} = \mathrm{arg}\min_{{u}^{i}\in\mathbb{R}^m}\|{u}^{i}-{u}_{\ell}^{i}(o^{i})\|^2 \mathrm{~\st{}~}({u}^{i}-\bar{{u}}_{d}^{i})^{\top}e^{i}_v \leq 0.
\end{align}
where $\bar{{u}}_{d}^{i}$ is given by~\eqref{def_us_bar_lagrange}. Suppose that there exists a control input ${u}^{i}$ that satisfies the constraint of~\eqref{eqopt_lag} for each $i$. The safety of the system~\eqref{eqlagrange} is then guaranteed when $d^{i}=0$ and $\Gamma^{i}=0$, \ie{}, all the agents will not collide with the other objects when there is no external disturbance.

Also, the problem~\eqref{eqopt_lag} is always feasible and its solution is given by ${u}^{i}_{\mathrm{opt}}={u}_{d}^{i}(o^{i})$ for ${u}_{d}^{i}$ of~\eqref{safe_control}, thereby minimizing the deviation of the safe control input from ${u}_{\ell}^{i}(o^{i})$.
\end{theorem}
\begin{proof}
Let us consider the following Lyapunov-type function:
\begin{align}
    \mathcal{V}_{s} = k_{p}\psi(X)+\sum_{i=1}^N\frac{\|v^{i}-v^{i}_{d}\|_{M^{i}(p^{i})}^2}{2}
\end{align}
where $v^{i}_{d}$ is given in~\eqref{safe_control} and $\psi$ is given as
\begin{align}
    \label{def_psi_all}
    \psi(X) = -\log\left\{\left(\prod_{i=1}^N\prod_{\substack{j>i\\j\in\mathscr{N}^{i}}}h(p^{ij})\right)\left(\prod_{i=1}^N\prod_{j\in\mathscr{M}^{i}}h(p^{ij})\right)\right\}~~~~
\end{align}
for $X = \{x^{1},\cdots,x^{N}\}$. By the definition of $h$ in~\eqref{safety_def}, the safe operation of the system~\eqref{eqlagrange} is guaranteed as long as $\psi^{i}$ of~\eqref{safety_def} is bounded. Taking the time derivative of $\mathcal{V}_{s}$, we get
\begin{align}
    \dot{\mathcal{V}}_{s} = \sum_{i=1}^N{k_{p}\nabla_{p^{i}}\psi^{i}}^{\top}v^{i}_{d}-k_{v}\|e^{i}_v\|_{M^{i}}^2+({u}^{i}-\bar{{u}}_{d}^{i})^{\top}e^{i}_v \label{last_eq}
\end{align}
by using~\eqref{eqlagrange} for $d^{i}=0$ and $\Gamma^{i}=0$ with the relation $z^{\top}(\dot{M}^{i}-2C^{i})z=0,~\forall z\in\mathbb{R}^n$, where $e^{i}_v=v^{i}-v^{i}_{d}$ and the arguments are omitted. Having $({u}^{i}-\bar{{u}}_{d}^{i})^{\top}e^{i}_v \leq 0$ as in the constraint of~\eqref{eqopt_lag} gives $\dot{\mathcal{V}}_{s}\leq \sum_{i=1}^N-{k_{p}}^2\|\nabla_{p^{i}}\psi^{i}\|^2-k_{v}\|e^{i}_v\|_{M^{i}}^2$, which guarantees the boundedness of $V$ and then $\psi^{i}$ for all $i$, implying no safety violation as long as the system is initially safe. Also, the constraint $({u}^{i}-\bar{{u}}_{d}^{i})^{\top}e^{i}_v\leq0$ is always feasible for ${u}^{i}$ given as ${u}^{i}=\bar{{u}}_{d}^{i}$. Finally, applying the KKT condition~\cite[pp. 243-244]{citeulike:163662} to~\eqref{eqopt_lag} results in ${u}^{i}_{\mathrm{opt}}={u}_{d}^{i}(o^{i})$ for ${u}_{d}^{i}(o^{i})$ given in~\eqref{safe_control}.
\end{proof}
\subsection{Optimal Safety Filter for General Nonlinear Systems}
\label{sec_optimal_safety_general}
This section generalizes the results of Theorem~\ref{thm_lagrangian_safe} for nonlinear systems given by~\eqref{equnderactuated}.
\subsubsection{Incremental Lyapunov Function}
\label{incremental_lyapunov}
We use the following incremental Lyapunov function inspired by contraction theory~\cite{Ref:contraction1,tutorial}, which leads to a safety analysis analogous to that of LTV systems and Lagrangian systems~\eqref{eqlagrange}:
\begin{align}
    &\mathcal{E}^{i}(o^{i},t) = \frac{\|v^{i}-v^{i}_{d}(o^{i})\|_{M^{i}(x^{i},t)}^2}{2} \label{simple_lyapunov} \\
    &\text{\st{}~}\dot{M}^{i}+M^{i}A_{d}^{i}+{A_{d}^{i}}^{\top}M^{i}-2M^{i}B^{i}{R^{i}}^{-1}{B^{i}}^{\top}M^{i} \leq -k_{v}M^{i} \label{simple_contraction}
\end{align}
where $M^{i}(x^{i},t)\succ0$, $k_{v}$ is a design parameter, and $A_{d}^{i}(o^{i},t)$ is a nonlinear state-dependent coefficient matrix for the system of~\eqref{equnderactuated} defined as follows:
\begin{align}
    \label{sdc_matrix}
    A_{d}^{i}(o^{i},t)(v^{i}-v^{i}_{d}(o^{i})) = f^{i}(p^{i},v^{i},t)-f^{i}(p^{i},v^{i}_{d}(o^{i}),t).
\end{align}
The arguments are omitted in~\eqref{simple_contraction} and the notations~\eqref{simple_lyapunov}~and~\eqref{simple_contraction} are intentionally consistent with the ones of Theorem~\ref{thm_lagrangian_safe} to imply the analogy between the methods in Sec.~\ref{sec_optimal_safety} and Sec.~\ref{sec_optimal_safety_general}. Note that the nonlinear matrix $A_{d}$ always exists when $f^{i}$ is continuously differentiable~\cite{nscm}.

The underlying benefit of using contraction theory here is that the problem of finding $M^{i}$ in~\eqref{simple_contraction} can be expressed as a convex optimization problem for optimal disturbance attenuation~\cite{mypaperTAC}. Although we use one of the simple versions of a contraction metric in this paper for simplicity of discussion, we can consider more general types of Lyapunov functions and contraction metrics for the sake of the broader applicability of our approach.
Reviewing how to find $M^{i}$ in~\eqref{simple_contraction} is beyond the scope of this paper, but those interested in knowing more about it can refer to~\cite{tutorial,chuchu_survey} and references therein.
\subsubsection{Augmenting General Learned Policy with Safety}
Let us first introduce the following assumption for generalizing the result of Theorem~\ref{thm_lagrangian_safe}.
\begin{assumption}
\label{assump_natural}
Consider a Lyapunov-type function for~\eqref{equnderactuated} defined as follows:
\begin{align}
    \mathcal{V}_{s}^{i}(o^{i},t) = k_{p}\psi^{i}(o^{i})+\frac{\|v^{i}-v^{i}_{d}(o^{i})\|_{M^{i}(x^{i},t)}^2}{2}
\end{align}
for $k_{p}>0$, $\psi^{i}(o^{i})$ of~\eqref{safety_def}, $v^{i}_{d}(o^{i})$ of~\eqref{safe_velocity}, and $M^{i}(x^{i},t)$ of~\eqref{simple_contraction}. We assume that
\begin{align}
    {e}^{i}_v &= B(p^{i},v^{i},t)^{\top}\nabla_{v^{i}} \mathcal{V}_{s}^{i} = B(p^{i},v^{i},t)^{\top}M^{i}(v^{i}-v^{i}_{d}) = 0 \label{ev_general_def} \\
    &\Rightarrow\dot{\mathcal{V}}_{s}^{i} \leq -{k_{p}}^2\|\nabla_{p^{i}}\psi^{i}(o^{i})\|^2-k_{v}\mathcal{E}^{i}(o^{i},t) \label{safety_condition_i}
\end{align}
where $e^{i}_v=v^{i}-v^{i}_{d}$ for $v^{i}_{d}$ of~\eqref{safe_velocity}, $\psi^{i}(o^{i})$ is given in~\eqref{safety_def}, $\alpha$ is given in~\eqref{safe_velocity}, $\mathcal{E}^{i}(o^{i},t)$ is given in~\eqref{simple_lyapunov}, and the arguments of $\mathcal{V}_{s}$ and $M^{i}$ are omitted for notational simplicity.
\end{assumption}

This assumption simply says that the system naturally satisfies the safety condition~\eqref{safety_condition_i} when the velocity displacements are in the directions orthogonal to the span of the actuated directions as discussed in~\cite{ccm}.
\begin{remark}
\label{remark_assumption}
Assumption~\ref{assump_natural} always holds for fully and over-actuated systems \st{} $B(p^{i},v^{i},t)B(p^{i},v^{i},t)^{\dagger} = \mathrm{I}_{n\times n}$, where $B(p^{i},v^{i},t)^{\dagger}$ the Moore-Penrose pseudo inverse. 
Furthermore, even when the system is under-actuated and Assumption~\ref{assump_natural} does not hold, the error associated with the under-actuation can be treated robustly as to be seen in Sec.~\ref{sec_robustness}.
\end{remark}

Given the learned motion planning policy $u_{\ell}^{i}(o^{i})$ of Sec.~\ref{sec_learned_policy} for the system~\eqref{equnderactuated}, we design a control policy ${u}_{d}^{i}$ processed by our CaRT safety filter as~\eqref{safe_control}, where ${e}^{i}_v$ is now given by~\eqref{ev_general_def} and $\bar{u}_{d}^{i}$ is defined as
\begin{align}
\label{ubars_general}
    \bar{u}_{d}^{i }= 
    \begin{cases}
    \frac{{e}^{i}_v({e^{i}_v}^{\top}M^{i}(\dot{v}^{i}_{d}-f^{i}_{d})-k_{p}{e^{i}_v}^{\top}\nabla_{p^{i}}\psi^{i})}{\|{e}^{i}_v\|^2} & \text{if $\|{e}^{i}_v\| \neq 0$}\\
    0 & \text{otherwise}
    \end{cases}
\end{align}
with $f^{i}_{d}=f^{i}(p^{i},v^{i}_{d},t)$ and $k_{p}$ being a design parameter. The controller~\eqref{safe_control} is well-defined with the division by $\|{e}^{i}_v\|$ under Assumption~\ref{assump_natural}, because the relation $(u_{\ell}^{i}(o^{i})-\bar{u}_{d}^{i})^{\top}{e}^{i}_v = 0 \leq 0$ holds when $\|{e}^{i}_v\| = 0$.
\begin{theorem}
\label{thm_general_safe}
Consider the optimization problem~\eqref{eqopt_lag}, where $u_{\ell}^{i}$ is now given by the learned motion planning policy for~\eqref{def_motion_planning} with~\eqref{equnderactuated}, ${e}^{i}_v$ is by~\eqref{ev_general_def}, $u_{\ell}^{i}$ is by~\eqref{ubars_general}.
Suppose that Assumption~\ref{assump_natural} holds and that there exists a control input $u^{i}$ that satisfies the constraint of~\eqref{eqopt_lag} for each $i$. The safety of the system~\eqref{equnderactuated} is then guaranteed when $d^{i}=0$ and $\Gamma^{i}=0$, \ie{}, all the agents will not collide with the other objects when there is no external disturbance.

Also, the problem~\eqref{eqopt_lag} is always feasible and its solution is given by ${u}^{i}_{\mathrm{opt}}={u}_{d}^{i}(o^{i})$ for ${u}_{d}^{i}$ of~\eqref{safe_control}, thereby minimizing the deviation of the safe control input from ${u}_{\ell}^{i}(o^{i})$.
\end{theorem}
\begin{proof}
Let us consider a Lyapunov-type function $\mathcal{V}_{s} = \psi(X)+\sum_{i=1}^N\mathcal{E}^{i}(o^{i},t)$, where $\psi$ is given in~\eqref{def_psi_all}, $v^{i}_{d}$ is given in~\eqref{safe_control}, and $\mathcal{E}^{i}(o^{i},t)$ is given in~\eqref{simple_lyapunov}. Using the relation~\eqref{simple_contraction} and Assupmption~\ref{assump_natural}, we have $\dot{\mathcal{V}}_{s} \leq \sum_{i=1}^N-{k_{p}}^2\|\nabla_{p^{i}}\psi^{i}\|^2-k_{v}\mathcal{E}^{i}(o^{i},t) +(u^{i}-\bar{u}_{d}^{i})^{\top}{e}^{i}_v$
when $d^{i}=0$ and $\Gamma^{i}=0$ in~\eqref{equnderactuated} as in the proof of Theorem~\ref{thm_lagrangian_safe}, where $\psi^{i}(o^{i})$ is given in~\eqref{safety_def}, $\alpha$ is given in~\eqref{safe_velocity}, and $\mathcal{E}^{i}(o^{i},t)$ is given in~\eqref{simple_lyapunov}. The rest follows from the proof of Theorem~\ref{thm_lagrangian_safe} below~\eqref{last_eq}.
\end{proof}
\begin{remark}
The safety filter of Theorem~\ref{thm_general_safe} minimizes the deviation of the safe control input from the learned motion planning input of the general system~\eqref{equnderactuated}, which implies that it instantaneously minimizes the contribution of the under-actuation error when Assumption~\ref{assump_natural} does not hold. This error can be then treated robustly as discussed in Remark~\ref{remark_assumption}.
\end{remark}
\section{Robustness to External Disturbance}
\label{sec_robustness}
The results in Sec.~\ref{sec_safety} depend on the assumption that $d^{i}=0$ and $\Gamma^{i}=0$ in~\eqref{eqlagrange}~and~\eqref{equnderactuated}. This section discusses the safety of these systems in the presence of external disturbance.
\subsection{Revisiting Built-in Robustness of Safety Filter}
\label{sec_built_in_robustness}
Due to its Lyapunov-type formulation in Theorems~\ref{thm_lagrangian_safe}~and~\ref{thm_general_safe}, our safety filter inherits the
robustness properties discussed in, \eg{},~\cite{Khalil:1173048}. It also inherits the robustness of the barrier function of~\cite{robust_cbf,cbf_clf,high_order_cbf1} (\cite{sto_stability_book,stochastic_barrier2} for stochastic disturbance) as seen in Sec.~\ref{sec_learned_policy} with Fig.~\ref{concept_fig1} and~\ref{concept_fig2}.

This section is for hierarchically augmenting such built-in robustness with the tracking-based robustness as in the tube-based motion planning~\cite{tube_nmpc,hiro_tube} to lighten the burden of the safety filter in dealing with the disturbance. Given a safety condition ${h}\geq 0$, these two ways of augmenting the learned motion planning with robustness as in Fig.~\ref{concept_fig1} are achieved by
\begin{enumerate}[label={\color{caltechpurple}{(\alph*)}}]
    \item changing our safety filter parameters (\eg{}, making $k_p$ and $k_v$ larger in Theorems~\ref{thm_lagrangian_safe}~and~\ref{thm_general_safe}) \label{robust_1}
    \item tracking a safe trajectory that satisfies ${h} \geq 0$, ensuring the deviation from the perturbed trajectory is finite. \label{robust_2}
\end{enumerate}
The first approach~\ref{robust_1} could lead to a large repelling force due to the stability of the safe set originating from the use of $\dot{{h}}$, especially when we use the log-barrier formulation with the dynamics discretization (\ie{}, we get a larger control input as the agents get closer to the safety boundary, implying a large discretization error). In contrast, \ref{robust_2} does not involve such behavior as it simply attempts to track the safe trajectory satisfying ${h}\geq 0$. As illustrated in Fig.~\ref{concept_fig2}, there are the following two sources of robustness in our approach:
\begin{enumerate}[label={\color{caltechred}{(\alph*)}}]
    \item asymptotic/exponential stability of the safe set~\label{robust_1_summary}
    \item incremental asymptotic/exponential stability of the system trajectory with respect to a safe target trajectory~\label{robust_2_summary}
\end{enumerate}
and this section is about~\ref{robust_2_summary}, which significantly reduces the responsibility of the safety filter~\ref{robust_1_summary} in meeting the robustness requirement, allowing the safe set to be less stable (\ie{}, the unsafe set to be less repelling, meaning more freedom in choosing the safety filter parameters). These observations will be more appreciable in the numerical simulations in Sec.~\ref{sec_simulation}.
\subsection{Optimal Robust Filter for Lagrangian Systems}
\label{sec_robustness_proof}
Let us consider the following control policy~${u}_{r}^{i}(o^{i},t)$ for the CaRT robust filter of the Lagrangian system~\eqref{eqlagrange}:
\begin{align}
    \label{robust_control}
    {u}_{r}^{i}(o^{i},t) =  {u}_{d}^{i}(t)-\begin{cases} 
   \frac{s^{i}({u}_{d}^{i}(t)-\bar{{u}}_{r}^{i})^{\top}s^{i}}{\|s^{i}\|^2} & \text{if $({u}_{d}^{i}(t)-\bar{{u}}_{r}^{i})^{\top}s^{i} > 0$} \\ 
    0 & \text{otherwise}
    \end{cases}~~~~
\end{align}
where $s^{i} = (v^{i}-v^{i}_d(t))+\Lambda_r^{i}(p^{i}-p^{i}_d(t))$ for a positive definite position control gain $\Lambda_r^{i}\succ 0$, $p^{i}_d(t)$, $v^{i}_d(t)$, and ${u}_{d}^{i}(t)$ are the safe target position, velocity, and control input computed by integrating~\eqref{eqlagrange} with the safe control input~\eqref{safe_control} assuming $d^{i}=0$ and $\Gamma^{i}=0$, respectively, and $\bar{{u}}_{r}^{i}$ is given as
\begin{align}
    \bar{{u}}_{r}^{i}=M^{i}\dot{s}^{i}+C^{i}s^{i}+G^{i}+D^{i}-k_r^{i}M^{i}s^{i}
\end{align}
with its arguments omitted for simplicity and $k_r^{i}>0$ being a scaler control gain for the composite state error $s^{i}$. We have the following result. Note that $x_d^{i} = [{p_d^{i}}^{\top},{v_d^{i}}^{\top}]^{\top}$ corresponds to $x_d$ of the conceptual illustration in Fig.~\ref{concept_fig1}.
\begin{theorem}
\label{thm_robust_lagrange}
Suppose that the system~\eqref{eqlagrange} is controlled by~\eqref{robust_control} and that the target safe trajectory of each agent is expressed with $p^{i}_d(t)$, $v^{i}_d(t)$, and ${u}_{d}^{i}(t)$ of~\eqref{robust_control}. If there exist bounded positive constants $\underline{m}^{i}$, $\overline{m}^{i}$, $\bar{d}_s^{i}$, $\overline{m}^{i}_{x}$, and $\overline{m}^{i}_{x^2}$ satisfying $\underline{m}^{i}\mathbb{I}\preceq M^{i}\preceq\overline{m}^{i}\mathbb{I}$, $\|(M^{i})^{-1}\Gamma^{i}\|_F^2 \leq \bar{d}_s^{i}$,  $\|\partial M^{i}/\partial x_k\| \leq \overline{m}^{i}_{x}$, and $\left\|{\partial^2M^{i}}/{(\partial x_k\partial x_{\ell})}\right\| \leq \overline{m}^{i}_{x^2},~\forall x,t$, then there exists an appropriate set of the control parameters and the positive scalar $\Delta r_s$ of~\eqref{safety_def}, which guarantee a safe operation of the system~\eqref{eqlagrange} at time $t$ with a finite probability, even under the presence of external disturbance.

Also, the controller \eqref{robust_control} is the optimal solution to
\begin{align}
    \label{eqopt_lag_robust}
    &{u}^{i}_{\mathrm{opt}} = \mathrm{arg}\min_{{u}^{i}\in\mathbb{R}^m}\|{u}^{i}-{u}_d^{i}(t)\|^2 \mathrm{~\st{}~}({u}^{i}-\bar{{u}}_{r}^{i})^{\top}s^{i} \leq 0.
\end{align}
\ie{}, ${u}^{i}_{\mathrm{opt}}={u}_{r}^{i}(o^{i},t)$, which is always feasible with $({u}^{i}-\bar{{u}}_{r}^{i})^{\top}s^{i} \leq 0$ representing an incremental exponential stability condition, and thus it minimizes the deviation of the robust control input from the learned safe control input ${u}_d^{i}(t)$.
\end{theorem}
\begin{proof}
The first part follows from the incremental stability analysis with the Lyapunov function $ \mathcal{V}_{r}^{i} = {s^{i}}^{T}M(p^{i})s^{i}$. Applying the weak infinitesimal operator $\mathscr{A}$~\cite[p. 9]{sto_stability_book} for analyzing the time evolution of~\eqref{eqlagrange}, we get
\begin{align}
    \mathscr{A} \mathcal{V}_{r}^{i} \leq -(k_r^{i}-{\underline{m}^{i}}^{-1}((\bar{d}^{i}+\bar{d}_s^{i}\overline{m}^{i}_{x}){\varepsilon_d^{i}}^{-1}+\bar{d}_s^{i}\overline{m}^{i}_{x^2}/2)) \mathcal{V}_{r}^{i}+\underline{m}^{i}C_d^{i}\nonumber
\end{align}
where $\bar{d}^{i}$ is given in~\eqref{eqlagrange}, $\varepsilon_d^{i}>0$ is a design parameter, and $C_d^{i} = (\bar{d}_s^{i}\overline{m}^{i}+\varepsilon_d^{i}(\bar{d}^{i}+\bar{d}_s^{i}\overline{m}^{i}_{x}))/\underline{m}^{i}$. If we select $k_r^{i}$ and $\varepsilon_d^{i}$ to have $2\bar{k}_r^{i} \leq (k_r^{i}-{\underline{m}^{i}}^{-1}((\bar{d}^{i}+\bar{d}_s^{i}\overline{m}^{i}_{x}){\varepsilon_d^{i}}^{-1}+\bar{d}_s^{i}\overline{m}^{i}_{x^2}/2))$ for some $\bar{k}_r^{i} > 0$, then the application Dynkin's formula~\cite[p. 10]{sto_stability_book} and the Gronwall-type lemma~\cite{gronwall} gives
\begin{align}
    \mathbb{E}[\|s^{i}\|] \leq \sqrt{\frac{C_d^{i}}{2\bar{k}_r^{i}}}+\sqrt{\left[\mathbb{E}[ \mathcal{V}_{r}^{i}|_{t=0}]-\frac{C_d^{i}}{2\bar{k}_r^{i}}\right]^{+}}e^{-\bar{k}_r^{i}t} = a^{i}+b^{i}e^{-\bar{k}_r^{i}t}\nonumber
\end{align}
where $[\cdot]^{+} = \max\{0,\cdot\}$. The hierarchical structure of the controller~\eqref{robust_control} along with Markov's inequality~\cite[pp. 311-312]{probbook} yields $\mathbb{P}[\|p^{i}(t)-p_d^{i}(t)\| > D_s^{i}] \leq D_{\mathbb{E}}^{i}/D_s^{i}$, where $D_s^{i} > 0$, $D_{\mathbb{E}}^{i} =\mathbb{E}[\|e^{i}_d(0)\|^2]e^{-\underline{\lambda}_r^{i}t}+{a^{i}(1-e^{-\underline{\lambda}_r^{i}t})}/{\underline{\lambda}_r^{i}}+{b^{i}(e^{-\bar{k}_r^{i}t}-e^{-\underline{\lambda}_r^{i}t})}/{(\underline{\lambda}_r^{i}-\bar{k}_r^{i})}$, and $\underline{\lambda}_r^{i}\mathbb{I}\preceq\Lambda_r^{i}$. Since $p_d(t)$ is a safe trajectory, selecting $\Delta r_s$ of~\eqref{safety_def} as $\Delta r_s=D_s^{i}$ guarantees safety at time $t$ with probability $[1-D_{\mathbb{E}}^{i}/D_s^{i}]^{+}$.

The optimality and feasibility argument follows as in the proof of Theorem~\ref{thm_lagrangian_safe}.
\end{proof}
\begin{remark}
The safety argument of Theorem~\ref{thm_robust_lagrange} is based on the assumption that the target trajectory is safe for all $t$. Therefore, each agent updates the safe target trajectory as in~\cite{7989693} if new obstacles/agents are reported or their goal region is changed. We can also incorporate the uncertainty associated with this fact by modifying the minimal safe distance $r_s$ of~\eqref{safety_def} larger than the actual one, or by formulating and solving stochastic motion planning problems in training the learned motion planning policy~${u}_{\ell}^{i}(o^{i})$ of Sec.~\ref{sec_learned_policy}. If the disturbance and the uncertainty are too large to be treated robustly, then we can further generalize our approach using nonlinear adaptive control and system identification for nonlinear systems~(see, \eg{}, Sec.~8~and~9 of~\cite{tutorial}.).
\end{remark}
\subsection{Optimal Robust Filter for General Nonlinear Systems}
\label{sec_robustness_proof_general}
Due to the existence of the matrix function $M^{i}$ in~\eqref{simple_contraction}, we can readily obtain a robustness result of Theorem~\ref{thm_robust_lagrange} even for general nonlinear systems~\eqref{equnderactuated}, simply by replacing each ${u}$ of~\eqref{robust_control} with $u$ as we derived the results of Theorem~\ref{thm_general_safe} analogously to Theorem~\ref{thm_lagrangian_safe}. We thus omit the derivation of the robustness guarantee in this paper due to space limitation without repeating the argument, but those interested can refer to the existing literature such as~\cite{cdc_ncm,mypaperTAC,tutorial} which explicitly discusses the error bound as the one of Theorem~\ref{thm_robust_lagrange}.
\section{Numerical Simulation}
\label{sec_simulation}
This section demonstrates the effectiveness of CaRT for several motion planning and control problems. 
\subsection{Illustrative Examples}
\label{sec_examples}
\subsubsection{General Nonlinear System}
\label{sec_nonlienar_example}
Let us first consider a nonlinear dynamical system~\eqref{equnderactuated} with $f^{i}$ given as\textcolor{white}{~\eqref{example_dynamics}}
\begin{align}
    f^{i}(p^{i},v^{i},t) = 
    \begin{bmatrix}
    \cos(p_1^{i})p_2^{i}-v_1^{i}+v_2^{i} \\
    -\sin(p_2^{i})p_1^{i}v_2^{i}+(v_1^{i})^2-v_2^{i}-2v_1^{i}v_2^{i}
    \end{bmatrix} 
    \label{example_dynamics}
\end{align}
where $N=1$ (number of agents) and $M=5$ (number of obstacles). Our safety filter and robust filter are constructed using Theorem~\ref{thm_general_safe} and the methods outlined in Sec.~\ref{sec_robustness_proof_general}, where the state-dependent coefficient matrix is used to construct the contraction metric of~\eqref{simple_contraction} as in~\cite{mypaperTAC,tutorial}. This example does not have a clear physical interpretation, but finding the incremental Lyapunov function is non-trivial as the system is nonlinear and non-polynomial and the set of CLFs is non-convex~\cite{ccm}.

\begin{figure}
    \centering
    \includegraphics[width=85mm]{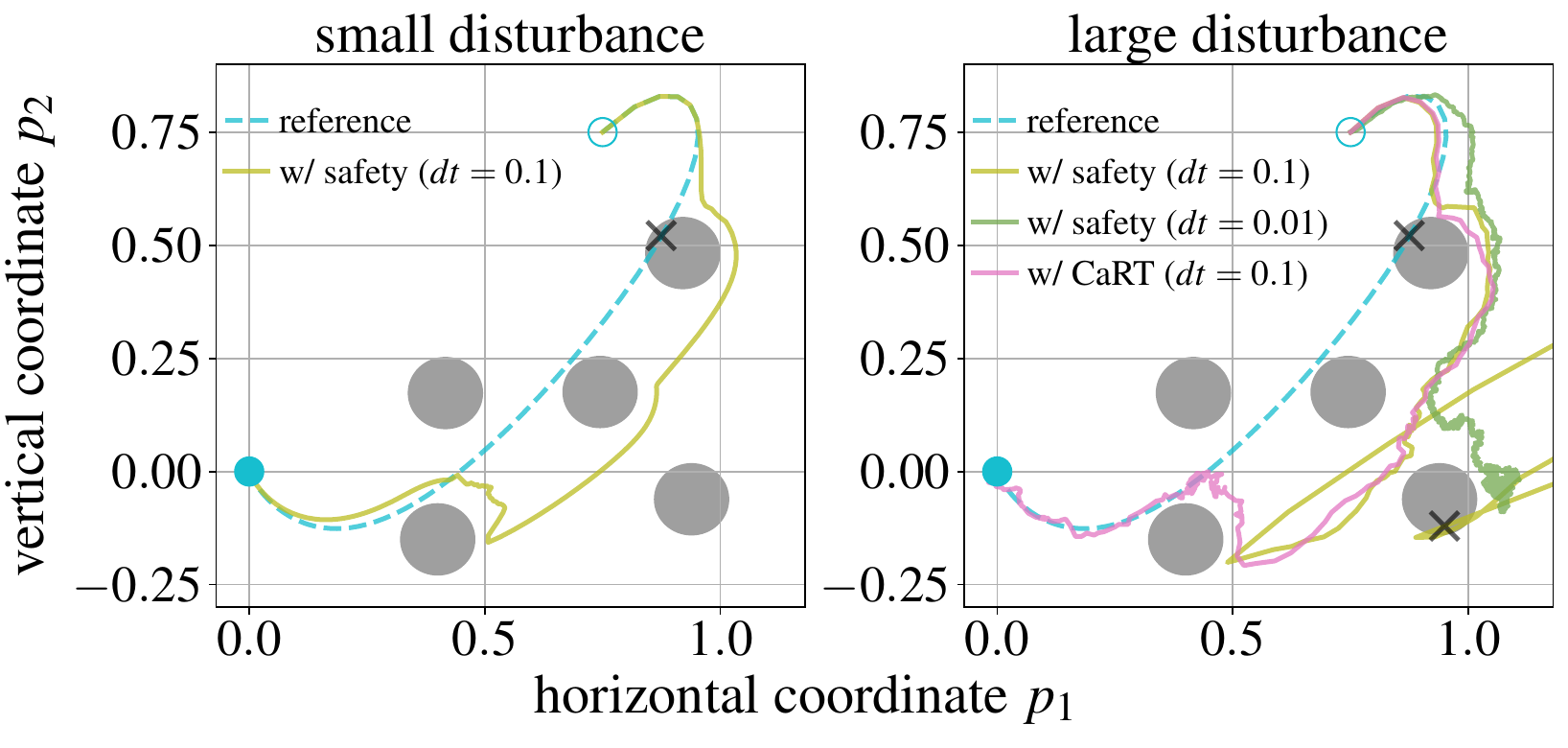}
    \vspace{-1.2em}
    \caption{Position trajectories of the nonlinear system~(\ref{equnderactuated}) with~(\ref{example_dynamics}) with small and large disturbance, $\bar{d}^{i}= \bar{\gamma}^{i}=2.0\times 10^{-5}$ and $\bar{d}^{i}= \bar{\gamma}^{i}=2.0\times 10^{-2}$ in~(\ref{equnderactuated}), respectively, which capture the trade-off discussed in~\ref{item_safety_robust} and~\ref{item_robust_robust}~of~Sec.~\ref{sec_trade_off}.}
    \label{nonilenar_fig}
\vspace{-1.3em}
\end{figure}

The reference trajectory shown in the left-hand side of Fig.~\ref{nonilenar_fig} is the one generated simply by a baseline contraction-based tracking control law~\cite{mypaperTAC,tutorial} with the target trajectory being the stationary origin $x_d(t) =[0,0,0,0]^{\top},~\forall t$. The initial condition is selected to be $x(0) = [1.0,1.0,1.0,1.0]^{\top}$. Because of the lack of safety consideration, it violates a safety constraint at the position indicated by $\times$. When the disturbance is small ($\bar{d}^{i}= \bar{\gamma}^{i}=2.0\times 10^{-5}$), our baseline safety filter works without the robust filter thanks to its built-in robustness even with the control interval $dt=0.1$, as we discussed in~\ref{item_safety_robust}~of~Sec.~\ref{sec_trade_off}. This is as expected from the argument of Sec.~\ref{sec_clfcbf}.

When $\bar{d}$ and $\bar{\gamma}$ get larger ($\bar{d}^{i}= \bar{\gamma}^{i}=2.0\times 10^{-2}$) as in the right-hand side of Fig.~\ref{nonilenar_fig}, however, the baseline safety filter becomes too sensitive to the disturbance for the control interval $dt=0.1$, leading to a large control input and safety violation indicated by $\times$. This situation can be avoided by using a smaller control interval $dt=0.01$, but still, its control input for safety becomes more dominant than the ideal control input for the reference trajectory even in this case, taking longer to reach the target position as can be seen from the green trajectory of Fig.~\ref{nonilenar_fig}. The combination of the safety filter and robust filter, CaRT, indeed allows for considering robustness separately when applying the safety filter, thereby handling large disturbances without violating safety and losing too much of the reference control performance, as discussed in~\ref{item_robust_robust}~of~Sec.~\ref{sec_trade_off}.
\subsubsection{Thruster-based Spacecraft Simulators}
\label{sec_scobs}
\begin{figure*}
    \centering
    \includegraphics[width=165mm]{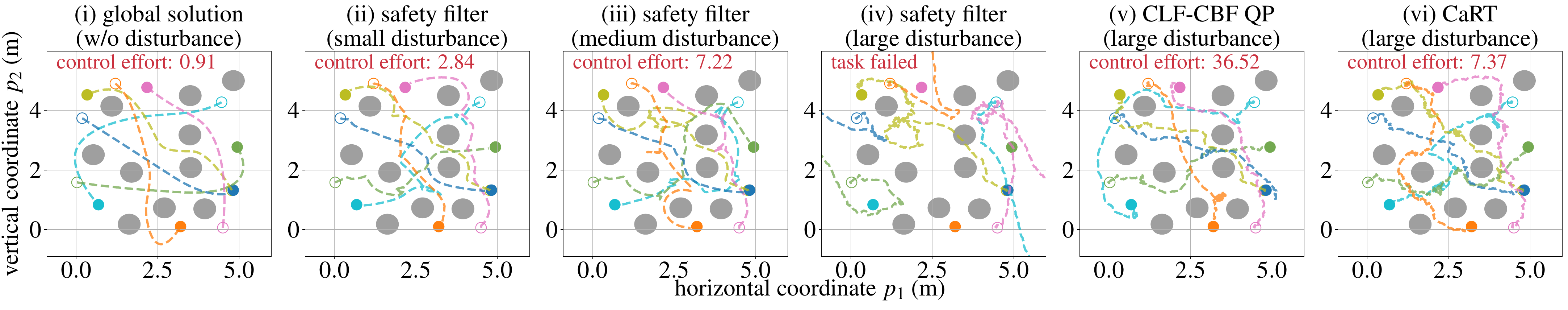}
    \vspace{-1.2em}
    \caption{Position trajectories of the spacecraft simulator system~\cite{SCsimulator} with various disturbances (small: $\bar{d}^{i}= \bar{\gamma}^{i}=3.0\times 10^{-4}$, medium: $\bar{d}^{i}= \bar{\gamma}^{i}=5.0\times 10^{-3}$, and large: $\bar{d}^{i}= \bar{\gamma}^{i}=5.0\times 10^{-2}$ in~(\ref{equnderactuated})), which capture the trade-off discussed in~Sec.~\ref{sec_high_level_overview}.}
    \label{scobs_fig}
\vspace{-1.5em}
\end{figure*}
Such a trade-off is more evident in a practical multi-agent robotic system, where we cannot use a smaller control time interval due to hardware limitations. We next consider a nonlinear spacecraft simulator system given in~\cite{SCsimulator} with $N=6$ (number of agents), $M=10$ (number of obstacles), and $r_{\mathrm{sen}} = 2.0$~(\si{\meter}) (sensing radius), where the control time interval $dt$ is required to be $dt \geq 0.1$~(\si{\second}). The dynamics parameters are normalized to $1$.

The learned motion planning policy detailed in Sec.~\ref{sec_learned_policy} is constructed using a neural network used in~\cite{lagros} with the training process outlined in~\cite{glas}. It utilizes the centralized global solution data sampled by solving~\eqref{def_motion_planning} using, \eg{}, the sequential convex programming, for the decentralized approximation by the neural network with the local observation~\eqref{observation}. The initial and target states of the spacecraft and the positions of the circular static obstacles in Fig.~\ref{scobs_fig} are randomized during the training and simulation. The cost function for the objective function is selected as $c(x(\tau),u(\tau),\tau)=\sum_{i=1}^N\|u^{i}(\tau)\|^2$ in~\eqref{def_motion_planning}.

As shown in Fig.~\ref{scobs_fig}, we can indeed observe the differences between each approach as discussed in Sec.~\ref{sec_high_level_overview} with Fig.~\ref{concept_fig1} and~\ref{concept_fig2}. 
When the size of the disturbance is relatively smaller than the learning error ((ii) of Fig.~\ref{scobs_fig}), then our baseline safety filter works with its built-in robustness (see~\ref{item_safety_robust} of Sec.~\ref{sec_trade_off}). The loss of optimality in control results from the presence of disturbance and its decentralized implementation with distributed information. 
As the disturbance gets larger ((iii)~--~(v) of Fig.~\ref{scobs_fig}), the baseline safety filter starts to fail. Also, the CLF-CBF approach starts to yield excessively large control input even when its QP is solved with a control input constraint ($|(u^{i})_k| \leq 1.00$, $k=1,\cdots,n$), which results in additional computational burden for each agent (see~\ref{clfcbf_item1}~--~\ref{clfcbf_item3} of Sec.~\ref{sec_clfcbf} with Fig~\ref{concept_fig2}). Task failure is defined as the situation where at least one of the spacecraft does not reach the target state.
Even when the size of the disturbance is relatively larger than the learning error ((vi) of Fig.~\ref{scobs_fig}), CaRT, the safety filter equipped with the robust filter, still works, retaining its control effort $4.96$ times smaller than that of the CLF-CBF approach and $8.10$ times greater than that of the optimal solution (see~\ref{item_robust_robust} of Sec.~\ref{sec_trade_off} with Fig.~\ref{concept_fig1} and~\ref{concept_fig2}).
\subsection{Multi-Spacecraft Reconfiguration}
\label{sec_multi_sim}
\begin{figure}
    \centering
    \includegraphics[width=85mm]{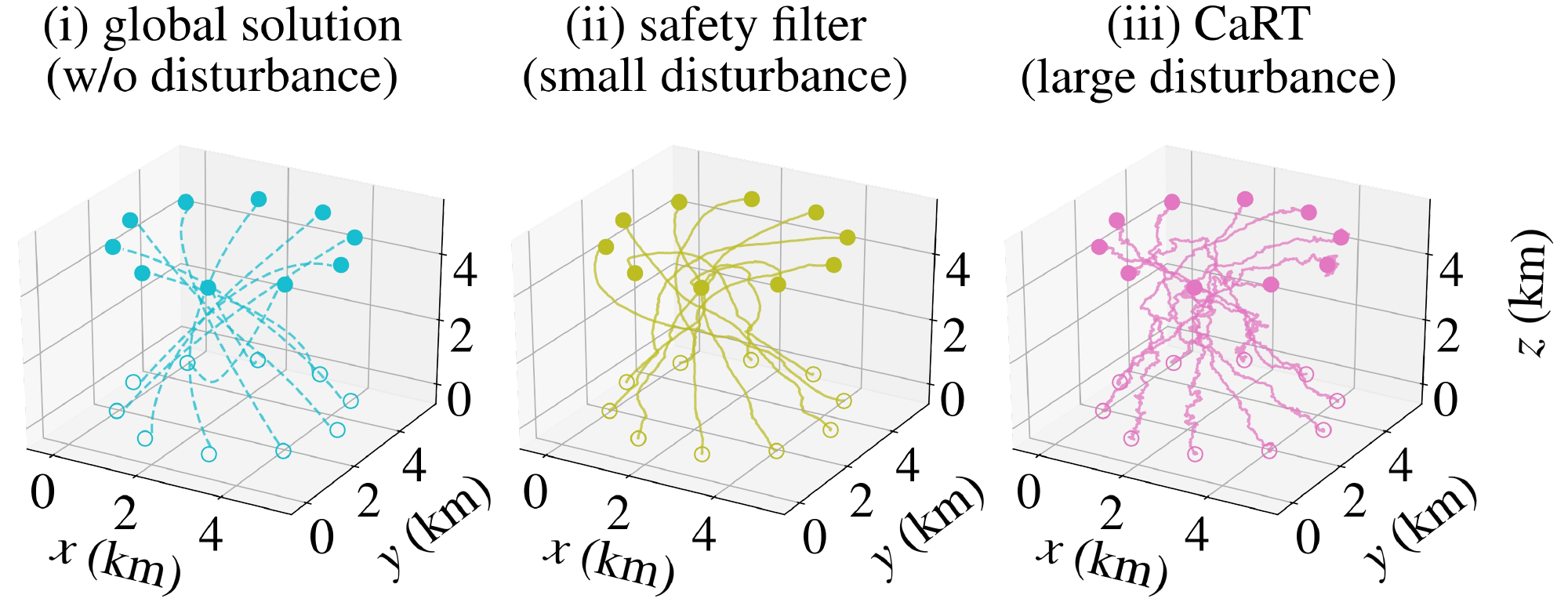}
    \vspace{-1.2em}
    \caption{Position trajectories for the multi-spacecraft reconfiguration task in LEO~\cite{doi:10.2514/1.37261} with various disturbances (small: $\bar{d}^{i}= \bar{\gamma}^{i}=1.0\times 10^{-2}$ and large: $\bar{d}^{i}= \bar{\gamma}^{i}=5.0\times 10^{-2}$ in~(\ref{eqlagrange})).}
    \label{leo_fig}
\vspace{-0.7em}
\end{figure}
As a part of JPL's CASTOR project, let us consider the optimal reconfiguration in Low Earth Orbit (LEO) with $10$ spacecraft ($N=10$ and $M=0$). The distributed motion planning policy of Sec.~\ref{sec_learned_policy} is trained again using~\cite{glas} as discussed in Sec.~\ref{sec_scobs} with $r_{\mathrm{sen}} = 2.0$~(\si{\meter}) for the sensing radius. Its dynamical system can be expressed as a Lagrangian system~\eqref{eqlagrange} as in~\cite{doi:10.2514/1.37261}. Task success is defined as the situation where the agent safety reaches a given target terminal state $x_f$ within a given time horizon. The success rate is computed as the percentage of successful trials in the total $50$ simulations. The initial and target states are randomized during the training and simulation. Also, the cost function of~\eqref{def_motion_planning} is again selected as $c(x(\tau),u(\tau),\tau)=\sum_{i=1}^N\|u^{i}(\tau)\|^2$ in~\eqref{def_motion_planning}. 

As implied in Fig.~\ref{leo_fig} and as discussed in Sec.~\ref{sec_scobs}, we can still see that the baseline safety filter works for small disturbances, and CaRT, the safety filter equipped with the robust filter, works for large disturbances. Such an observation can be corroborated by the results summarized in Table~\ref{tab_leo}. In particular, CaRT augments the learned motion planning policy with safety and robust tracking, resulting in its success rate of \SI{100}{\percent} with its control effort $4.64$ times greater than that of the optimal solution. Again, the loss of optimality in control results from the presence of disturbance and the CaRT's decentralized implementation with distributed information. 
\renewcommand{\arraystretch}{1.1}
\begin{table}
\caption{Control performances for the multi-spacecraft reconfiguration in LEO averaged over $50$ simulations, where $J=\int^{t_f}_0\sum_{i=1}^N\|u^{i}\|^2d\tau$. \label{tab_leo}}
\vspace{-2em}
\begin{center}
\begin{tabular}{|l|c|c|}
\hline
guidance and control methods  & success rate (\%) & control effort $J$ \\
\hline
\hline 
global solution (w/o disturb.) & $100$ & $1.26$ \\
safety filter (small disturb.) & $94.0$ & $4.93$ \\
CLF-CBF QP (lrg. disturb.)~\cite{cbf_clf} & $31.6$ & $10.36$ \\
CaRT (large disturb.) & $100$ & $5.86$ \\
\hline
\end{tabular}
\end{center}
\vspace{-0.6em}
\scriptsize{$^*$small disturb.: $\bar{d}^{i}= \bar{\gamma}^{i}=1.0\times 10^{-2}$ and large disturb.: $\bar{d}^{i}= \bar{\gamma}^{i}=5.0\times 10^{-2}$ in~(\ref{eqlagrange}).}
\vspace{-1.5em}
\end{table}
\renewcommand{\arraystretch}{1.0}
\section{Conclusion}
\label{sec_conclusion}


In this paper, we present CaRT, a hierarchical, control-theoretic filter for augmenting machine-learning motion planning algorithms with certified safety and robust tracking guarantees, for a large class of multi-agent nonlinear dynamical systems.
Unlike existing methods, CaRT uses a safety filter, which steers the system into the safe set, to certify the safety of the learned policy, then uses a robust filter, which steers the system into the safe trajectory, to reject deterministic and stochastic disturbances. 
As demonstrated in numerical experiments, this hierarchical nature allows CaRT to guarantee safety and robustness under much larger disturbances in off-nominal settings. This makes a major distinction from conventional safety-driven approaches including CLF-CBF, where the robustness could originate from the over-conservative pulling force to the interior of the safe set.

\appendices

\bibliographystyle{IEEEtran}
\bibliography{root}

\end{document}